\documentclass[journal]{IEEEtran}

\IEEEoverridecommandlockouts                              % This command is only needed if 
\usepackage{booktabs,balance,subfigure}
\usepackage{graphicx,balance}
\usepackage{multirow}
\usepackage[sort,compress]{cite}
\let\labelindent\relax
\usepackage{enumitem}
\usepackage{xcolor}
\usepackage{amsmath}
\setlist[itemize]{align=parleft,left=0pt..1em}

\usepackage[pdftex, pdfstartview={FitV},bookmarksopen=true,plainpages = false, colorlinks=true, linkcolor=black, citecolor = black, urlcolor = black,filecolor=black , pagebackref=false,hypertexnames=false, plainpages=false, pdfpagelabels ]{hyperref}

\title{\LARGE \bf 
Wide-Area Geolocalization with a Limited Field of View Camera in Challenging Urban Environments
}

\author{Lena M. Downes$^{1, 2, 3}$, Ted J. Steiner$^{2}$, Rebecca L. Russell$^{2}$ and Jonathan P. How$^{1}$% <-this % stops a space
\thanks{$^{1}$Department of Aeronautics and Astronautics,
        Massachusetts Institute of Technology, Cambridge, MA, USA}%
\thanks{$^{2}$ Perception and Embedded ML Group,
        Draper, Cambridge, MA, USA}%     
\thanks{$^{3}$ Draper Scholar. Research funded by Draper. {\tt\small lmdownes@mit.edu}}%
}

\begin{document}

\maketitle
\thispagestyle{empty}
\pagestyle{empty}

%%%%%%%%%%%%%%%%%%%%%%%%%%%%%%%%%%%%%%%%%%%%%%%%%%%%%%%%%%%%%%%%%%%%%%%%%%%%%%%%
\begin{abstract}
Cross-view geolocalization, a supplement or replacement for GPS, localizes an agent within a search area by matching ground-view images to overhead images. Significant progress has been made assuming a panoramic ground camera. Panoramic cameras' high complexity and cost make non-panoramic cameras more widely applicable, but also more challenging since they yield less scene overlap between ground and overhead images. This paper presents Restricted FOV Wide-Area Geolocalization (ReWAG), a cross-view geolocalization approach that combines a neural network and particle filter to globally localize a mobile agent with only odometry and a non-panoramic camera. ReWAG creates pose-aware embeddings and provides a strategy to incorporate particle pose into the Siamese network, improving localization accuracy by a factor of 100 compared to a vision transformer baseline. This extended work also presents ReWAG*, which improves upon ReWAG's generalization ability in previously unseen environments. ReWAG* repeatedly converges accurately on a dataset of images we have collected in Boston with a 72\textdegree{} field of view (FOV) camera, a location and FOV that ReWAG* was not trained on.
%ReWAG trains on image pairs with known camera heading and location within the satellite %image to produce embeddings that emphasize the area of the satellite image that is visible %in the ground image. At runtime ReWAG uses noisy compass heading and particle location %within the satellite image to generate pose-aware embeddings. 
% We include in our submission a video that demonstrates ReWAG's convergence on a test path of several dozen kilometers.
%A video highlight that demonstrates ReWAG's convergence on a test path of several dozen kilometers is available at \url{https://youtu.be/U_OBQrt8qCE}.

\end{abstract}
%%%%%%%%%%%%%%%%%%%%%%%%%%%%%%%%%%%%%%%%%%%%%%%%%%%%%%%%%%%%%%%%%%%%%%%%%%%%%%%%

\section{INTRODUCTION}
\begin{figure}[t!]
  \centering
  \includegraphics[width=\linewidth]{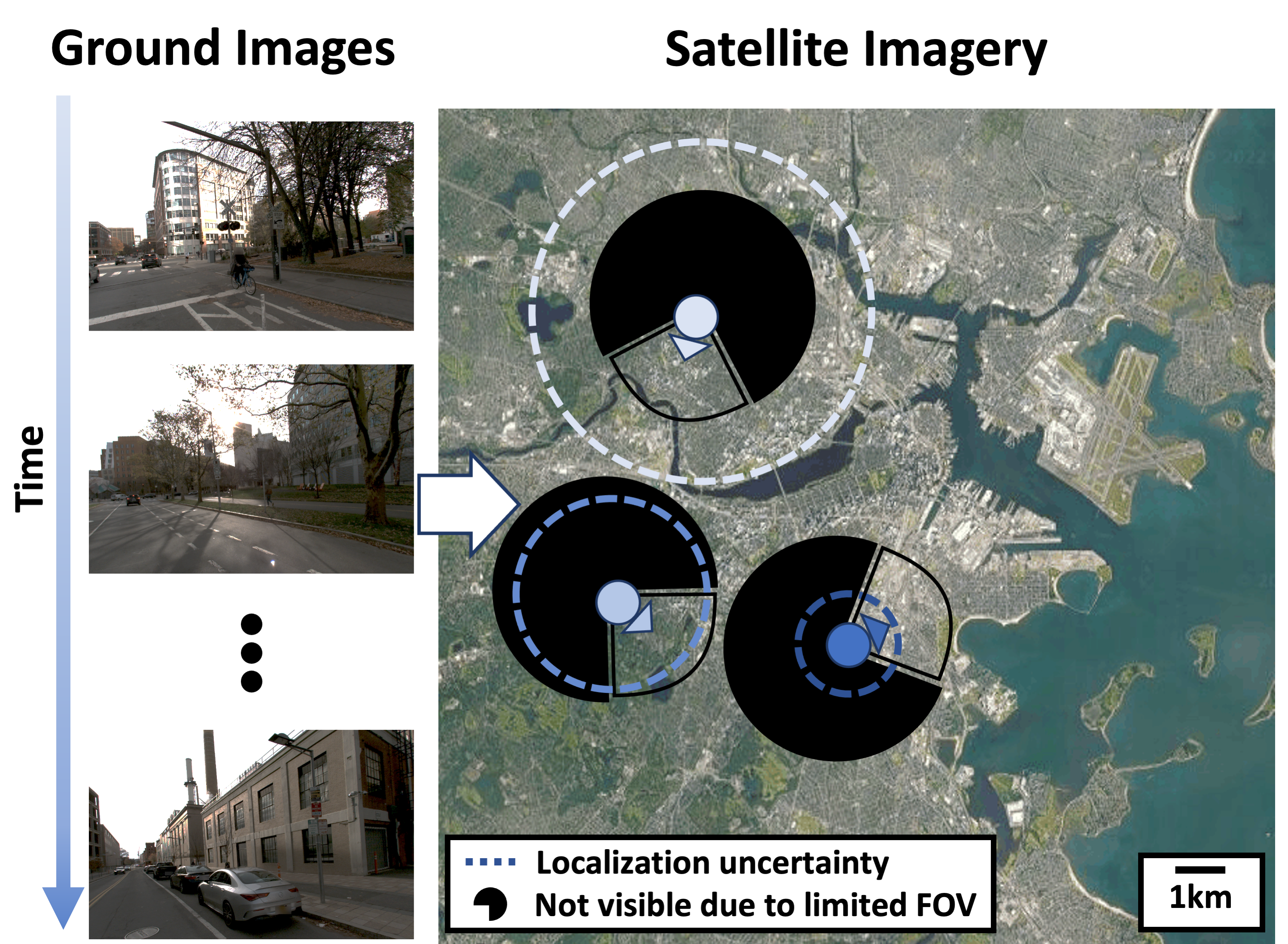}
    \vspace*{-0.25in}
  \caption{ReWAG is a cross-view geolocalization system that takes in a series of non-panoramic ground-view camera images and satellite imagery of the search area to accurately localize the agent on a search area scale that was not possible with previous works.}
  \label{fig:intro}\vspace*{-.2in}
\end{figure}

GPS is an external system for localization that is susceptible to failure through jamming, spoofing, and signal dropout due to dense foliage or urban canyons. Cross-view geolocalization \cite{Tian, Shi, Cai, Kim, Hu} is a localization method that only requires images from a ground-view camera and preexisting overhead imagery, with or without GPS measurements. Cross-view geolocalization measures the similarity between a ground image and all of the satellite images in a search area to determine the location that the ground image was taken from (see Fig.~\ref{fig:intro}). Satellite imagery at some resolution is widely available for most of the planet, even in forests and urban areas where GPS signals can be weaker.

Cross-view geolocalization with ground and overhead images is challenging due to the wide difference in viewpoints. Many existing works \cite{Hu, ZhuVIGOR, Liu2019, Viswanathan} rely upon the use of a panoramic ground camera because it effectively decreases the problem dimensionality by reducing the impact of heading. Although a panoramic ground camera can have different headings, the heading only affects the alignment of the image, not the content, whereas the heading of a limited field of view (FOV) camera affects the visible content of the image. The use of panoramic ground cameras also simplifies the problem by maintaining as much semantic similarity as possible between the two viewpoints---overhead images show 360\textdegree{} of the surroundings of a ground agent, as do panoramic ground cameras. However, in practice panoramic cameras are rarely used due to their high monetary cost (resulting in lesser availability) and their difficulty to mount without occlusion. As a result, few real-world systems can benefit from panoramic-based localization. Widespread adoption of cross-view geolocalization technology will require its applicability to platforms without panoramic imaging capabilities. 

Most recent work on cross-view geolocalization takes a deep learning approach to the problem by using Siamese networks \cite{Kim, Hu, Cai, Liu2019, Zhu2021, ZhuVIGOR}. A Siamese network consists of a pair of neural networks with matching architectures that simultaneously learn embedding schemes for ground and overhead images. The Siamese network is trained so that images taken in similar locations are close together in embedding space, and images taken at different locations are embedded far apart. 
The accuracy of the Siamese networks can be further improved by integrating their measurements over time with a particle filter \cite{Xia, ZhuVIGOR, Hu, Kim, downes_iros_22}. However, existing particle filter geolocalization works are highly constrained -- requiring some level of GPS data \cite{Xia, ZhuVIGOR}, 180\textdegree{} (or greater) FOV of the ground camera \cite{Xia, ZhuVIGOR, Hu, Kim, downes_iros_22}, perfect initial location knowledge \cite{Hu}, or a search area of less than 2.5 km$^2$ \cite{Kim, Hu}. These additional constraints are needed because the algorithms cannot efficiently geolocalize a limited FOV camera across a large search area. Localization with a limited FOV camera increases both the difficulty of the problem and the computational requirements.

\textbf{Our contribution.} Our approach, Restricted FOV Wide-Area Geolocalization (ReWAG), builds upon our previous work in \cite{downes_iros_22} to enable efficient wide-area geolocalization with a restricted FOV camera through two key changes. The first change is a computationally efficient method for matching limited FOV ground images to satellite images by appending relative pose information to the intermediate network embedding before inputting it to the Spatial Aware Feature Aggregator (SAFA) \cite{Shi}. The second change is the dual incorporation of relative pose into both the Siamese network and the particle filter, which enables the probability distribution to be modeled more accurately. The only additional information these changes necessitate at runtime is a noisy heading from a compass, and they produce a cross-view geolocalization system that is capable of localizing across city-scale search areas using a 90\textdegree{} FOV or less camera. In summary, in this paper we demonstrate the following contributions for restricted FOV cross-view localization: 
\begin{enumerate}[leftmargin=*] %,topsep=-1em
    \item Efficient generation of pose-aware embeddings that generalize to unseen environments;
    \item A particle filter system that more accurately models the probability distribution, resulting in lower average and final localization error;
    \item Faster particle filter convergence than a ViT baseline~\cite{zhu2022transgeo}; and
    \item A challenging new dataset for testing of urban cross-view geolocalization in Boston.
\end{enumerate}

\textbf{Differences from conference paper.} 
This evolved paper extends our previous work \cite{downes_icra_23} in three main ways. The first being that we have collected a dataset of images and GPS tags in Boston that were used to test ReWAG* and which is publicly available at \href{https://doi.org/10.5281/zenodo.7818704}{https://doi.org/10.5281/zenodo.7818704}. Testing on this dataset initiated the second aspect of the extension---changes in our particle filter resampling scheme. Namely, changing from multinomial resampling \cite{tibshirani1993introduction} at each time step to systematic resampling \cite{doucet2001sequential} when effective sample size \cite{liu1996metropolized} dips below a threshold. Testing on this Boston dataset also motivated the third aspect of the extension---incorporation of training data augmentations to improve performance on ground images with variable brightness. We retrained the Siamese network with ground image augmentations generated from Fancy PCA~\cite{krizhevsky2017imagenet}.

\section{RELATED WORKS}
\textbf{Ground-to-aerial cross-view geolocalization.} Cross-view geolocalization derives from previous work in the areas of scene recognition and image retrieval. Ground-to-aerial cross-view geolocalization pushes previous work to higher levels of difficulty due to the vast difference in viewpoints between ground and aerial images. Previous works have attempted to solve this problem with hand-crafted features and traditional computer vision techniques \cite{Viswanathan, viswanathan2016, jacobs, bansal}, but recent works \cite{Workman, Kim, Hu, Cai, Vo, Rodrigues, Shi, Cao, Zhu2021, Liu2019, lin, lin2015} have mostly applied deep learning in the form of Siamese networks \cite{bromley}. In recent years, recall at top-1 has been steadily rising, but most works \cite{Hu, ZhuVIGOR, Liu2019, Viswanathan} focus on panoramic ground images and report high recall at top-1 for these panoramic ground images. Recall at top-1 for limited FOV ground images is significantly lower than that for panoramic ground images. However, limited FOV cameras are much more common than panoramic cameras in practice. 
% Many works focus on image retrieval: determining the best matching satellite image to a ground image given a database of satellite images, where the performance metric is recall at top-k. Recall at top-k is the fraction of ground images for which the matching satellite image is ranked in the top-k of the satellite image database. 

% Recall at top-1 lies at around 50\% \cite{Zhu2021}, which means that image retrieval alone can only match a ground image to the correct satellite image 50\% of the time. With 50\% recall at top-1, we cannot trust that a single image retrieval gives us the correct satellite image match. One effective way to determine an accurate location estimate is to combine the information from many different image matches over time. Unlike those image retrieval focused systems, \sysname{} is designed to maximize the localization information gained from combining image matches over time.
\textbf{Orientation-aware cross-view geolocalization.}
Non-panoramic ground cameras make cross-view geolocalization more difficult due to the reduced number of visible features in the image and due to the matching satellite features being concentrated within one area of the satellite image instead of spread throughout it. The unknown orientation of the ground camera is a key factor in this problem. Some previous works have developed methods to incorporate an understanding of orientation into the cross-view geolocalization system, like by appending orientation maps to images to be input to the Siamese network \cite{Liu2019}, by using Dynamic Similarity Matching (DSM) to calculate the correlation between ground and satellite images \cite{shi2020looking}, or by jointly embedding the full satellite image as well as the satellite image portion that is visible in the limited FOV ground image \cite{rodrigues2022global}. Instead of jointly determining the most highly matching satellite image and the orientation, \cite{shi2022beyond} assumes that the satellite image has already been determined, and they then use pose optimization to estimate the pose within that satellite image. \textit{These works, not designed for mobile robotics constraints, require many search iterations, polar transformations and data augmentations at runtime and hence may be too computationally demanding for real-time robotics.}

\textbf{Orientation-blind cross-view geolocalization.}
More recent works tend to treat orientation as an aspect of the problem that can be solved at the last step \cite{Zhu2021}, or do not directly encode or estimate it at all \cite{Shi, zhu2022transgeo}. In \cite{Zhu2021} orientation-invariant embedding schemes are learned through a combination of global mining, binomial loss, and training data augmentation with random rotations. SAFA \cite{Shi} is an attention mechanism that helps the network to learn image descriptors regardless of the large viewpoint difference between ground and satellite views. TransGeo \cite{zhu2022transgeo} uses a vision transformer instead of the typical convolutional neural network (CNN) approach. This attention-focused approach embeds patches of the images into tokens with learnable position tokens, which gives it a more flexible method for learning about orientation and position while embedding images. \textit{Although orientation-blind embedding schemes can improve image retrieval when orientation is unknown, these methods do not have a mechanism by which the ground camera pose can be input when modeling with a particle filter.}

\begin{figure*}[t!]
\centering
  \includegraphics[width=0.9\linewidth]{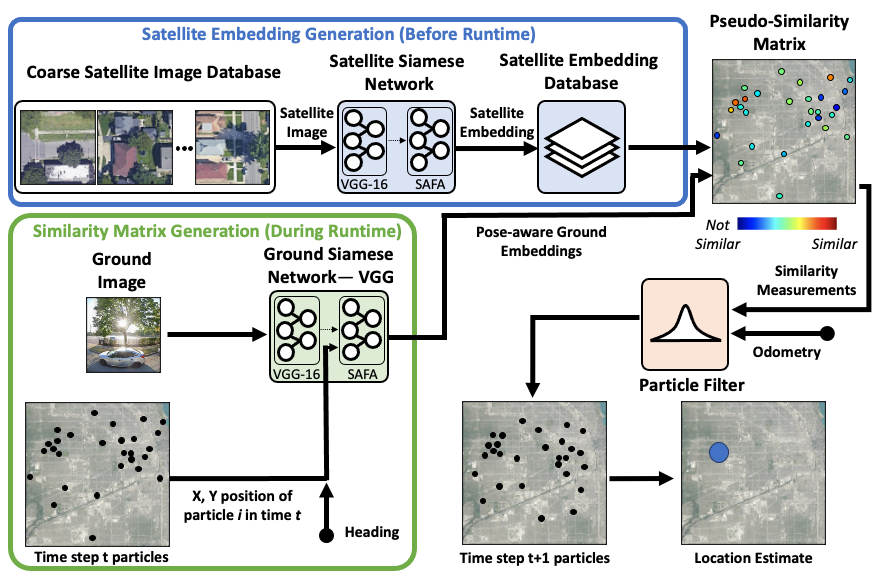}
  \caption{Diagram of ReWAG. Satellite embeddings are generated before runtime with the coarsely sampled satellite image database and the Siamese network that was trained with trinomial loss. During runtime, a pose-aware ground embedding is generated for each particle at each time step and combined with the satellite embeddings to create a pseudo-similarity matrix, which is the similarity of each particle with its location in the search area. Odometry and measurements from the pseudo-similarity matrix are input to the particle filter with a Gaussian measurement model to generate a location estimate.}
  \label{fig:system}
\vspace*{-0.2in}
\end{figure*}

\textbf{Particle filter implementations.} 
Some previous works have combined image retrieval with particle filters to enable cross-view geolocalization over time as an agent moves through a search area \cite{Kim,Hu,Xia, Viswanathan, viswanathan2016, downes_iros_22}. Particle filters in these systems use random discrete particles to model a probability distribution of the agent location given odometry, satellite images of the search area, and ground images. However, existing works have localized with particle filters that use wide angle ground cameras \cite{Hu, downes_iros_22, Viswanathan, viswanathan2016}.  Although \cite{Kim,Xia} localize a ground agent with a 180\textdegree{} ground camera instead of a panoramic 360\textdegree{} camera, commonly available cameras like those found in cell phones have FOV of 90\textdegree{}
 or less, making much fewer features visible in their images. \textit{These existing particle filter cross-view geolocalization systems are not able to accurately localize with extremely limited FOV images.}

\vspace{-0.1in}
\section{METHODS}
\subsection{Overview of Approach}
ReWAG builds upon WAG \cite{downes_iros_22} to enable localization across a wide search area with restricted FOV ground images. ReWAG uses WAG's strategies of creating a coarse satellite image database, generating embeddings with Siamese networks based on VGG-16 \cite{vgg} and SAFA \cite{Shi}, and localizing over time with a particle filter (see Fig.~\ref{fig:system}). However, ReWAG differs from WAG in two major ways: first, ReWAG generates pose-aware image embeddings in a computationally efficient manner, and second, ReWAG generates more informative similarity measures and hence more accurately models the agent location probability distribution by incorporating pose information from each particle into the Siamese network input. For non-panoramic ground imagery it is necessary to incorporate this additional information due to the increased difficulty of the problem. 

ReWAG first coarsely samples the search area to construct a database of satellite images that are preprocessed with a Siamese network before runtime to generate satellite embeddings. During runtime, a generic ground embedding is generated by the VGG-16 portion of the Siamese network for each ground image, and for each particle a pose-aware embedding is produced from the generic embedding and the particle's pose. The similarity between each particle's pose-aware ground embedding and its corresponding satellite embedding is calculated to produce the pseudo-similarity matrix, a probabilistic representation of the ground image's similarity with different areas of the search area. Then, the particle filter receives odometry and measurements from the pseudo-similarity matrix to produce a location estimate at each time step. ReWAG* is our extension to ReWAG beyond \cite{downes_icra_23}, which has the same structure and training as ReWAG except for modifications to training data augmentation and particle filter resampling. Now, we describe in detail the changes implemented in ReWAG to enable limited FOV localization and the changes in ReWAG* that improve its performance on challenging, realistic datasets.

\subsection{Pose-Aware Embeddings}
\begin{figure}[t!]
\centering
  \includegraphics[width=\linewidth]{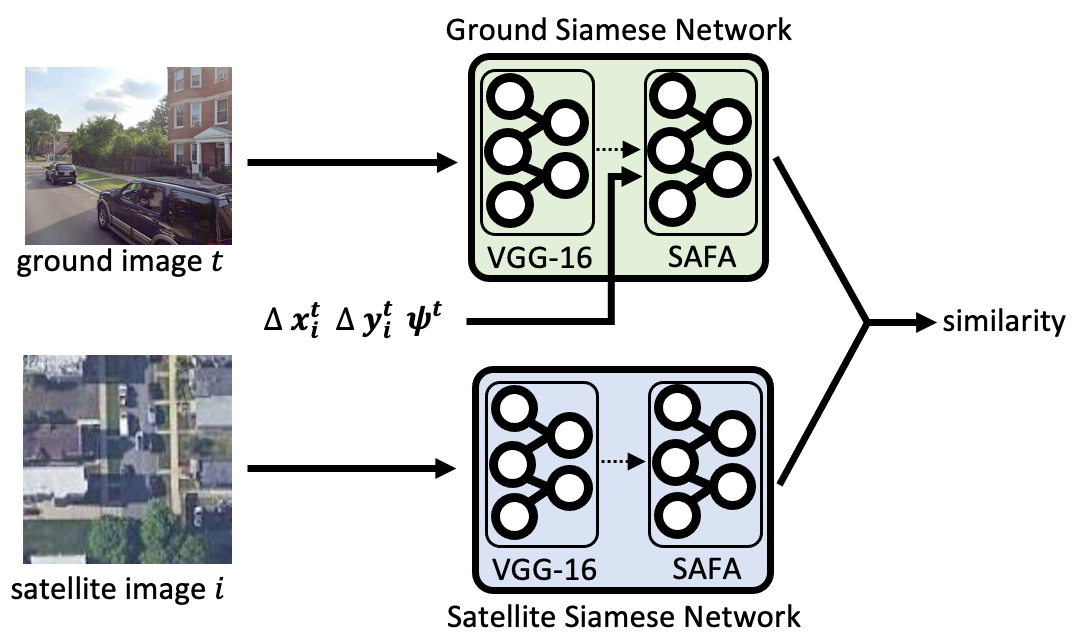}
%   \vspace*{-0.15in}
  \caption{For a given particle $i$ at time $t$, the ground image and the heading $\psi^t$ are given by the sensor measurements at time step $t$, the satellite image is given by the tile that particle $i$ is within, and the x and y displacements $\Delta x^t_i$ and $\Delta y^t_i$ are given by particle $i$'s location within its satellite tile.}
  \label{fig:pose_aware}\vspace*{-0.2in}
\end{figure}

We have developed a method to train the ground Siamese network to generate pose-aware embeddings in a computationally efficient manner while minimally modifying the architecture. Like WAG, our Siamese network architecture is derived from that of \cite{ZhuVIGOR}, which consists of a VGG-16 backbone and a SAFA module to increase the network's spatial understanding. In ReWAG, the ground Siamese network is modified to append the particle pose to the intermediate embedding that is output by the VGG-16 backbone, as shown in Fig.~\ref{fig:pose_aware}. This intermediate embedding appended with the particle pose is then input to SAFA, which learns a spatial-aware representation of the ground image. In terms of total computation, the VGG-16 backbone comprises 93\% and SAFA comprises 7\% of the time to produce one pose-aware embedding. The computationally efficient benefit of this method comes from the ability to generate one base embedding for each ground image, and then being able to append any pose to the base embedding to efficiently generate a pose-aware embedding with SAFA. When combined with a particle filter, this design enables the VGG-16 inference to be done once per time step instead of once for each particle for each time step. Then, the much lighter-weight SAFA is the only part that needs to do inference for each particle. In contrast, \cite{rodrigues2022global} generates pose-aware embeddings through a joint global and local pipeline, which means that the aerial image embedding can only be computed once heading is known. Hence there is no ability to generate aerial embeddings a priori, increasing computation at runtime.

\subsection{Training Data Augmentation} 
ReWAG* is additionally trained with Fancy PCA \cite{krizhevsky2017imagenet} data augmentations to improve its robustness to image brightness changes relative to ReWAG \cite{downes_icra_23}. Training data augmentation is a method to artificially increase the size of a neural network training dataset by applying transformations to the existing data. Augmentations often improve neural network robustness to changes that should not affect the neural network output. For example, image classifiers may be trained with random crops and rotations applied to the images, since these transforms do not affect the content of the image and the classifiers should predict the same class regardless. Since cross-view geolocalization is sensitive to spatial alignment, previous works have not incorporated data augmentation into the Siamese network training pipeline. Although spatial transform augmentations are not particularly helpful for cross-view geolocalization, brightness augmentations are. A Siamese network should be able to identify the matching satellite image for a given ground image no matter how dark or bright that ground image is, as long as the scene is still discernible. One method for brightness augmentation is Fancy Principal Component Analysis (Fancy PCA). Fancy PCA is a method for altering the intensities of pixel values in an image, done by adding random multiples of the image principal components to each image channel. Fancy PCA produces brightness changes with natural appearances. 

\subsection{Siamese Network and Particle Filter Integration}
\begin{figure}[t!]
\centering
  \includegraphics[width=\linewidth]{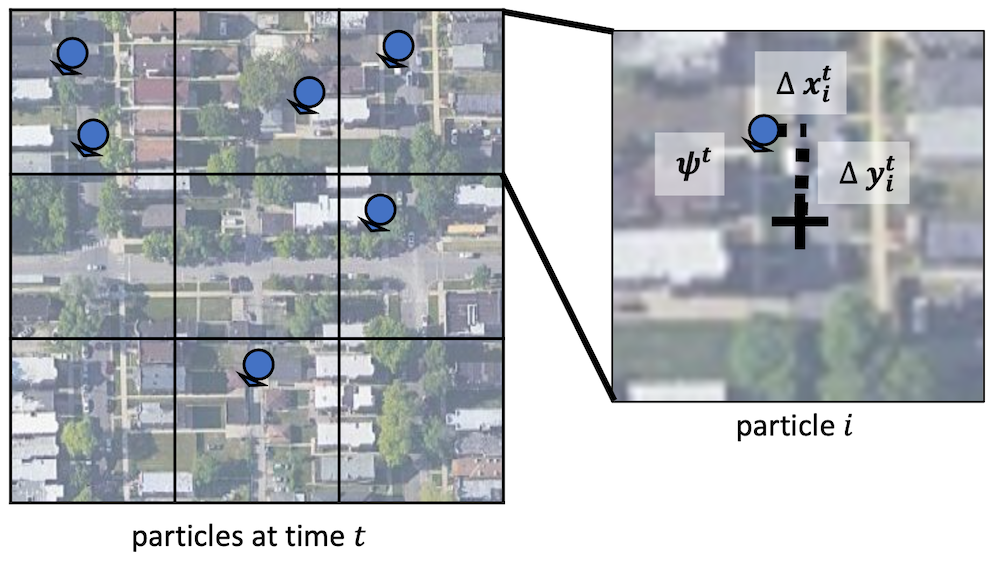}
%   \vspace*{-0.15in}
  \caption{Particles are dispersed through search area, which is segmented into satellite tiles whose embeddings are pre-computed before runtime. At each time step, the true heading of the ground agent is approximately known and the x, y location of each particle within its satellite tile is given by its displacement from the center of the tile. }
  \label{fig:particles}\vspace*{-0.2in}
\end{figure}

Our key observation is that cross-view geolocalization can be improved by more thorough integration between the Siamese network and the particle filter. Previous works have built systems with mostly one-way connections between the Siamese network and the particle filter---for each time step, the location of each particle determines which satellite image will be compared with that time step's ground image, and that similarity is used to adjust the weight of that particle. However, there is additional information that can be integrated into the Siamese network-particle filter connection. In addition to each particle having a corresponding satellite image, each particle also has a location within that satellite image, as shown in Fig.~\ref{fig:particles}. The location within the satellite image is part of the probability distribution that the particle filter is modeling, but in traditional architectures that information is not factored in to the similarity measure and hence it does not affect the particle weights. 

ReWAG includes a method for incorporating particle pose information into the similarity measure through the pose-aware embeddings. The pose of a particle $i$ at time $t$ consists of $x$ and $y$ displacements $\Delta x^t_i$ and $\Delta y^t_i$, which are determined from the particle's location within the associated satellite tile, and the heading $\psi^t$, which is determined from sensor measurements at time step $t$. We assume that the heading provided by the compass measurements will be accurate to the true ground agent heading within 2\% error. At each time step, the ground image is used to generate one generic intermediate embedding, and for each particle at that time step, a pose-aware embedding is generated from the generic embedding and each particle pose. Only one intermediate embedding is needed for each ground image because the image content is the same regardless of the particle pose. This method increases the computation required for each particle filter update compared to an orientation-blind approach, but the computationally efficient method by which we generate pose-aware embeddings helps to offset this increase. An orientation-blind approach only needs to generate one ground embedding at each time step, while an orientation-aware approach needs to generate one ground embedding for each particle at each time step. ReWAG is orientation-aware while needing only 7\% of the computation time required to generate one ground embedding for each particle at each time step, since it takes advantage of a shared intermediate embedding for each ground image. The incorporation of the particle pose aids the Siamese network in identifying where within the satellite image there should be corresponding features if the image pair is a positive match. The incorporation of domain-knowledge has been shown to improve deep neural network performance \cite{dash_review_2022}. Without pose-aware embeddings, the Siamese network has to learn where in the satellite image to look for matching features. With pose-aware embeddings, we tell the Siamese network where in the satellite image to look for matching features and hence simplify the learning task.

\subsection{Particle Filter Resampling} 
The evolved ReWAG* detailed in this work also benefits from improvements to its particle filter resampling strategy relative to \cite{downes_icra_23}.
Particle filter resampling is done after the reweighting step of the particle filter. Reweighting converts the weighted distribution of particles to an unweighted distribution by retaining and replicating particles with high weights, removing particles with low weights, and equalizing the particle weights. 

% Resampling algorithms first generate a cumulative sum of normalized weights
% \begin{equation}\label{eq:Qk}
% \begin{aligned}
% Q_k = \frac{W_k}{W_N} \qquad k \in 1{\dots} N \text{, with } W_{k} = {\sum }_{i=1}^{k}w_{i}
% \end{aligned}
% \end{equation}
% where $N$ is the number of particles and $w_i$ is the unnormalized weight of particle $i$. $N$ random numbers $u$ are then generated that are used to select particles for replication, which occurs for particle $x^k$ if:
% \begin{equation}\label{eq:ur}
% Q_{k-1} < u_r \leq Q_{k}, \qquad r \in 1{\dots} N
% \end{equation}

Different strategies for resampling determine the method by which particles are selected for replication or removal. %random numbers $u$ are generated. 
The most straightforward way to resample is 
multinomial resampling, as was done in \cite{downes_icra_23, downes_iros_22}. 
%
% In this process, the random numbers are generated according to
% \begin{equation}\label{eq:uniform}
% u_r = \tilde{u}_r \text{, with } \tilde{u}_r \sim \mathrm{U}[0,1) \qquad r \in 1{\dots} N.
% \end{equation}
This approach has the benefits of being relatively simple to implement, but it does a poor job of selecting particles uniformly relative to their weights. This property has been shown to result in poorer particle filter estimates \cite{hol}.
In ReWAG*, we have instead implemented systematic resampling as it has been shown to improve resampling quality and hence improve particle filter estimates over multinomial resampling \cite{hol}. 
% To perform systematic resampling, the random numbers are generated as
% \begin{equation}\label{eq:systematic}
% u_r = \frac{1}{N} (r-1) + \tilde{u}\text{, with } \tilde{u} \sim \mathrm{U}[0,\frac{1}{N}) \qquad r \in 1{\dots} N
% \end{equation}

We have also incorporated into ReWAG* effective sample size (ESS) \cite{Ristic2004BeyondTK}, a metric to turn the resampling function off and on. Resampling should only be done when new measurements are incorporated into the particle filter, because resampling without new measurements wastes computation and leads to sample impoverishment \cite{KUPTAMETEE2022110836}. In the bootstrap particle filter, which we previously used in ReWAG \cite{downes_icra_23} and WAG \cite{downes_iros_22}, resampling is done at each time step regardless of whether new measurements have been received. Our extended testing on data in Boston revealed particle degeneracy issues when performing the multinomial resampling done in \cite{downes_icra_23}, causing us to instead implement systematic resampling and to use ESS to control resampling frequency.
% One such criterion is when the effective sample size (ESS) drops below a specified percent of the total particles \cite{Ristic2004BeyondTK},
% \begin{equation}\label{eq:ess}
% ESS < \beta N \quad \text{where} \quad 
% ESS = \left(\sum _{i=1}^{N}{w_i^2}\right )^{-1},
% \end{equation}
% and $\beta$ is a parameter that determines the resampling sensitivity. 

% \begin{figure}[t!]
% \centering
%   \includegraphics[width=0.7\columnwidth]{sim_dist_transgeo.png}
%   \vspace*{-0.15in}
%   \caption{Baseline particle filter measurement model compared to empirical distribution of positive/semi-positive pair Euclidean distance measurements in the validation set from \cite{ZhuVIGOR}'s dataset. An exponential measurement model does not fit the data.}
%   \label{fig:base_meas_model}
% \end{figure}

% \begin{figure}[t!]
% \centering
%   \includegraphics[width=0.7\columnwidth]{sim_dist_mine.png}
%   \vspace*{-0.15    in}
%   \caption{\sysname{}'s particle filter measurement model of Eq.~\ref{eq:gauss} compared to empirical distribution in the validation set from \cite{ZhuVIGOR}'s dataset. \sysname{}'s measurement model better fits the data. Empirical data spike at 0 is due to some pairs being the database pair with maximum similarity.}
%   \label{fig:our_meas_model}\vspace*{-0.2in}
% \end{figure}

\section{RESULTS}
\subsection{Experimental Setup}
We use a version of the neural network architecture from \cite{ZhuVIGOR} with the VGG-16 backbone and the SAFA module, modified to generate pose-aware embeddings. The satellite network architecture is unchanged.  We train both ReWAG's Siamese network and our comparison baseline, TransGeo \cite{zhu2022transgeo}, on the VIGOR dataset \cite{ZhuVIGOR} with the ground images cropped to 90\textdegree{}. We train the TransGeo baseline for 50 epochs with the training parameters described in \cite{zhu2022transgeo}, and ReWAG for 30 epochs with triplet loss \cite{Hu}:
\begin{equation}\label{eq:triplet}
\mathcal{L}_{\text{triplet}} = \log\left(1+e^{-\alpha\left(d_{pos} - d_{neg}\right)}\right)
\end{equation}
with $\alpha$ loss parameter set to 10. Then we train ReWAG for 15 additional epochs with trinomial loss \cite{downes_iros_22}:
\begin{equation}\label{eq:trinomial}
\begin{aligned}
\mathcal{L}_{\text{t}} = \frac{\log\left(1+e^{-\alpha_{\text{p}}\left(S_{\text{p}}-m_{\text{p}}\right)}\right)}{N_{\text{p}}\alpha_{\text{p}}} +  \frac{\log\left(1+e^{\alpha_{\text{n}}\left(S_{\text{n}}-m_{\text{n}}\right)}\right)}{N_{\text{n}}\alpha_{\text{n}}} \\
+\frac{\log\left(1+e^{-\alpha_{\text{semi}}\left(S_{\text{semi}}-m_{\text{semi}}\right)}\right)}{N_{\text{semi}}\alpha_{\text{semi}}} 
\end{aligned}
\end{equation}

with the parameter values used in \cite{downes_iros_22}. The particle filter has 30,000 particles and uses the Gaussian measurement model from \cite{downes_iros_22}.

% \subsection{Chicago Simulation Setup}

% \begin{table}[t]
% \vskip+0.1in
% \caption{Trinomial Loss Parameters}
% \vspace*{-0.2in}
% \label{tab:loss_params}
% % \centering
% \begin{center}
% \begin{tabular}{|l|c|c|}\hline
% \bf Parameter & \bf Symbol & \bf Value \\\hline
% Positive pair weight &$\alpha_{p}$  & 5 \\%\hline
% Semi-positive pair weight &$\alpha_{s}$    & 6 \\%\hline
% Negative pair weight &$\alpha_{n}$    & 20 \\%\hline
% Positive pair average similarity &$m_{p}$  & 0 \\%\hline
% Semi-positive pair average similarity &$m_{s}$    & 0.3 \\%\hline
% Negative pair average similarity &$m_{n}$    & 0.7 \\\hline
% \end {tabular}
% % \centering
% \end{center}
% % \vspace*{-0.3in}
% \end{table}

% \end{table}

\subsection{Large-scale Simulation: Chicago} 
\begin{figure}[t!]
\centering
  \includegraphics[width=0.8\linewidth]{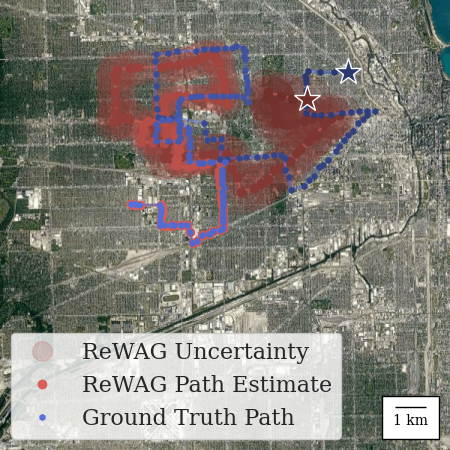}
%   \vskip-0.15in
  \caption{The ground-truth path in Chicago and ReWAG's path estimate, which accurately converges upon the ground-truth. Uncertainty bubble sizes are scaled down by raising to 0.75 to improve interpretability. Increasing brightness of path indicates passing of time. Start marked with stars.}
  \label{fig:path}
  \vskip-0.1in
\end{figure}

\textbf{Experiment details.}  To demonstrate ReWAG's performance relative to WAG \cite{downes_iros_22}, we perform limited FOV ground image localization experiments with the same simulated test paths from \cite{downes_iros_22} but, instead of using panoramic ground images, we crop the ground images to 90\textdegree{} FOV. The true path is shown in Fig.~\ref{fig:path} and is shown over our estimated location at each time step. These test paths are a large scale localization experiment with very noisy location initialization across the entire city of Chicago using simulated data. The simulated data consists of ground images and overhead satellite images from Google Maps Static API and odometry measurements from the ground-truth displacement between images with added noise proportional to displacement. A satellite image database was generated by sampling the search area approximately every 60 meters into a 256 $\times$ 256 grid of non-overlapping satellite image tiles. This grid size maintains a similar image size and resolution that the network was trained on; satellite images of size 640 $\times$ 640 pixels with a resolution of approximately 0.1 m/pixel. We initialized the particle filter with Gaussian distributions, centered 1.3 km from the true initial location (standard deviation of 900 m) for C-1 and C-2, and centered 600 m from the true initial location (standard deviation of 300 m) for C-3. We add 2\% noise to the ground-truth odometry and 1\% noise to the ground-truth heading at each time step. We ran this experiment with ReWAG and a baseline that uses TransGeo \cite{zhu2022transgeo}, a vision transformer-based approach, for its Siamese network combined with the same particle filter as ReWAG.

%   \vspace*{-0.15in}

\begin{figure}[t!]
  \centering
  \subfigure[Initial distribution supplied to particle filter. True location is over 1 km from initial particle filter estimate. \label{fig:initial_gauss}]{\includegraphics[width=.49\columnwidth]{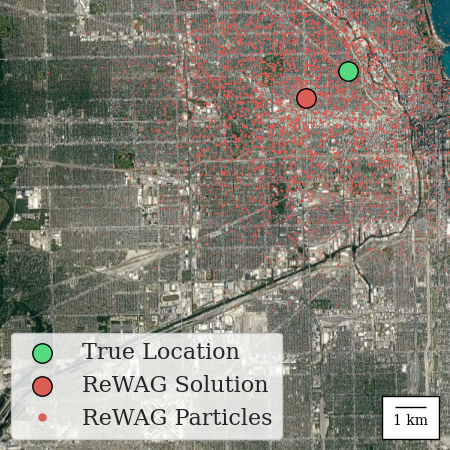}}
  \subfigure[Final particle filter solution and true location. True location is approximately 26 m from final particle filter estimate. \label{fig:zoom_nopart}]{\includegraphics[width=.49\columnwidth]{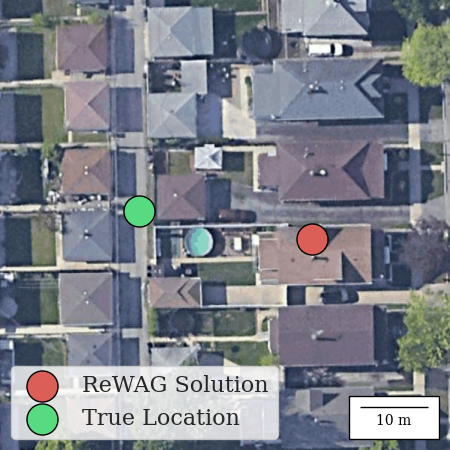}}
  \caption{ReWAG is able to accurately localize an agent to within 26 m of its true location across nearly 200 km$^2$ of Chicago after being initialized to a Gaussian distribution centered 1.3 km from the true location.}
%   \vspace*{-0.2in}
  \label{fig:bam}
\end{figure}

\begin{figure}[t]
\centering
  \includegraphics[width=0.85\linewidth]{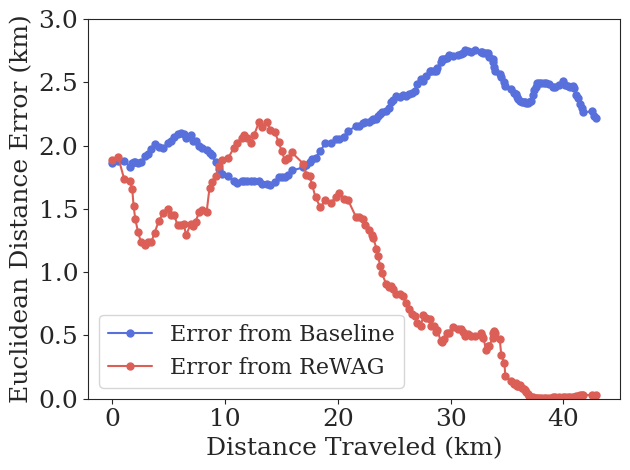}
  \vspace*{-0.2in}
\caption{Particle filter estimation error from ReWAG compared to a TransGeo baseline. ReWAG has lower final and average error.}
  \label{fig:err_conv}
%\end{figure}
% \vspace*{.1in}
%\begin{figure}[t]
%  \vspace*{-0.1in}
\centering
  \includegraphics[width=0.85\linewidth]{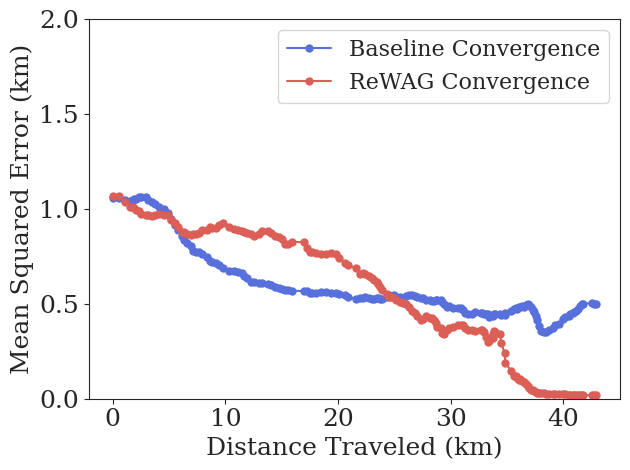}
  \vspace*{-0.2in}
\caption{Particle filter estimation convergence from ReWAG compared to a TransGeo baseline. ReWAG accurately converges, the baseline does not.}
  \label{fig:mse_conv}%\vspace*{-0.1in}
\end{figure}

\textbf{Estimation error.} Even with an initialization as far from the agent as Fig.~\ref{fig:initial_gauss}, ReWAG has a final estimation error of 26 m (visible in Fig.~\ref{fig:zoom_nopart}). This estimation error is only 5 meters greater than that which was achieved with 360\textdegree{} ground images in WAG. Fig.~\ref{fig:err_conv} compares the estimation error of ReWAG's particle filter and the TransGeo baseline as the simulated agent moves. The error is the Euclidean distance between the actual location and the weighted average of the locations of the particles in the particle filter.  Over the duration of the experiment, ReWAG has an average estimation error of 925 m, versus 2.2 km for the baseline. ReWAG achieves a final estimation error of 26 m compared to the baseline of 2.2 km. 

% \begin{figure}[t]
% %  \vspace*{-0.1in}
% \centering
%   \includegraphics[width=0.8\linewidth]{convergence_km_dist_std_icra.png}
%   \vspace*{-0.2in}
% \caption{Particle filter estimation convergence from ReWAG compared to baseline with TransGeo. ReWAG converges to an accurate estimate but the baseline does not.}
%   \label{fig:mse_conv}\vspace*{-0.1in}
% \end{figure}

\begin{figure}[t]
\centering
  \includegraphics[width=0.8\linewidth]{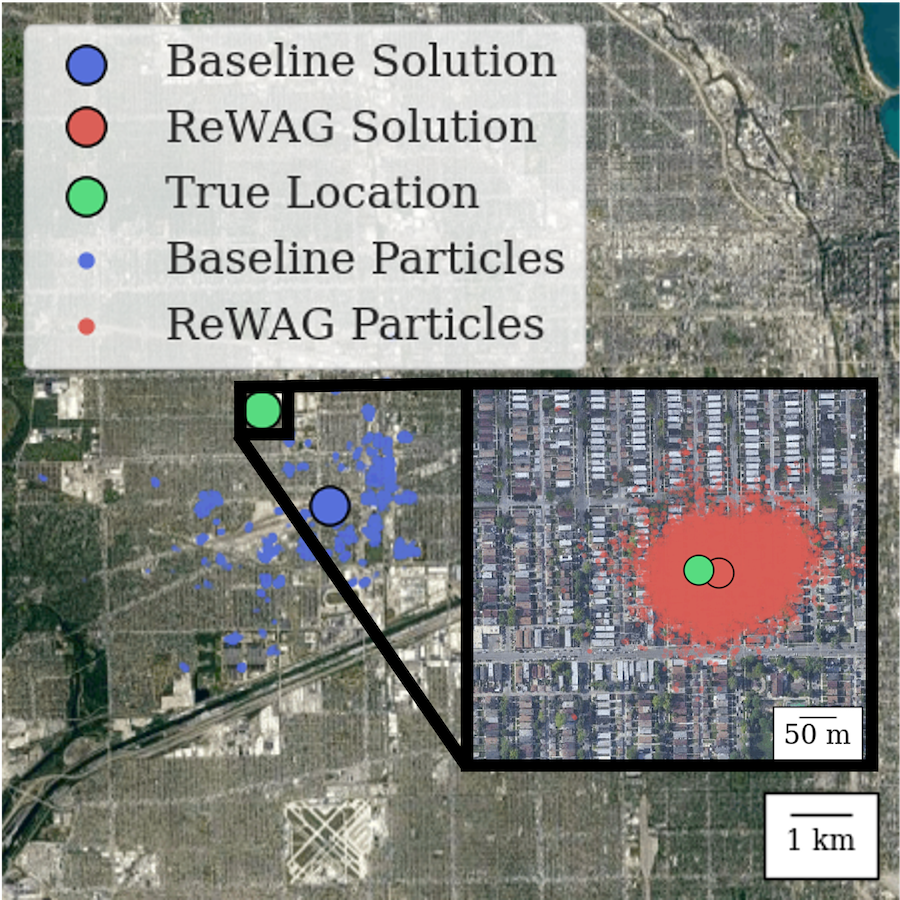}
  \vspace*{-0.1in}
\caption{Final particle filter dispersion with the baseline system and with ReWAG on the Chicago test path (C-1). Baseline does not successfully converge to a location estimate, while ReWAG converges to within 18 m of standard deviation.}
  \label{fig:exp_pf}
  \vspace*{.1in}
%\end{figure}
%\begin{figure}[t!]
\centering
  \includegraphics[width=0.8\linewidth]{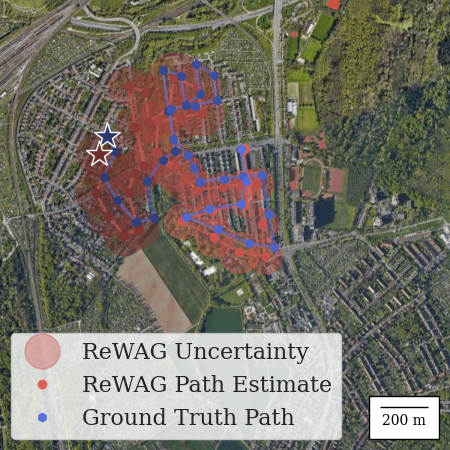}
\vspace*{-0.1in}
\caption{Ground-truth and estimated path in the KITTI test area. Increasing brightness of path indicates passing of time.  Start marked with stars.}
  \label{fig:kitti_path} 
\end{figure}

\textbf{Convergence.} Figure~\ref{fig:mse_conv} compares the system convergence measured as the mean squared error of the particle locations at each time step. ReWAG converges to a mean squared error of less than 60 m (the satellite image size) after 133 filter updates, while the baseline does not reach that level of convergence in the time period tested. Fig.~\ref{fig:exp_pf} shows the particle filter distribution that ReWAG converges to as red dots in the inset of the figure while the baseline terminates with the particle distribution shown in blue dots, which still shows significant estimation error.

\textbf{Ablation.} We performed a small ablation study to determine the benefit of including both heading and location information in the pose-aware embeddings as opposed to only including heading, or including neither as in WAG. We trained a Siamese network with the same architecture, training regime and parameters as ReWAG, with the exception that the base embeddings input to SAFA only had heading appended to them. We also tested limited FOV images with WAG. We tested the systems on the C-1 test path and the results are summarized in Table \ref{tab:summary_ablation}.
\begin{table}[t]
\caption{ReWAG Ablation}
\vspace*{-.2in}
\label{tab:summary_ablation}
\begin{center}
\begin{tabular}{|cl|c|c|c|}
\hline
\multicolumn{2}{|c|}{\bf Metric and System Type}  & \bf C-1  \\\hline
\multirow{3}{*}{Final Error (m)} & ReWAG & 26\\
& ReWAG without Position & 375 \\
& WAG & 192\\\hline
\multirow{3}{*}{Convergence Time (time steps)} & ReWAG & 133 \\
& ReWAG without Position & - \\
& WAG  & - \\\hline
\end{tabular}
\end{center} \vspace*{-0.1in}
\end{table} 

\setlength{\textfloatsep}{5pt}

% \vspace*{-0.15in}

% \textbf{Trinomial loss ablation.} Fig.~\ref{fig:vig_conv} shows an ablation study comparing ReWAG with and without trinomial loss training. ReWAG receives 30 epochs of training with triplet loss and 15 epochs with trinomial loss, while ReWAG without trinomial loss receives only 30 epochs of training with triplet loss. Both systems use the same particle filter. Although trinomial loss causes higher average error throughout the test path, it yields lower final estimation error once converged. The additional training with the trinomial loss improves image retrieval performance with semi-positive image pairs and contributes towards improved particle filter performance.

% \begin{figure}[t!]
% \centering
%   \includegraphics[width=0.8\linewidth]{convergence_km_dist_err_ablation_icra.png}
% \vspace*{-0.2in}
%   \caption{The estimation error on the same test path using without additional trinomial loss training.}
%   \label{fig:vig_conv}
%   \vspace*{-.2in}
% \end{figure}

\textbf{Multiple path result summary.} Table \ref{tab:summary_results} shows a summary of ReWAG's localization performance compared to the baseline on three test paths in Chicago. These paths are the same test paths that were used in \cite{downes_iros_22}. The first, C-1, is the path discussed previously (Fig.~\ref{fig:path}). The particle filter for C-1 and C-2 was initialized 1.2 km away from the ground truth, and for C-3 was initialized 600 m away from the ground truth due to the challenging urban scenery in this path. On all paths, ReWAG outperforms the baseline in final estimation error, final standard deviation and convergence time. 
\begin{table}[t]
\caption{Comparison of Results on Chicago Test Paths--\\ Baseline: With TransGeo}
\vspace*{-.2in}
\label{tab:summary_results}
\begin{center}
\begin{tabular}{|cl|c|c|c|}
\hline
\multicolumn{2}{|c|}{\bf Metric and System Type}  & \bf C-1  & \bf C-2 & \bf C-3 \\\hline
\multirow{2}{*}{Final Error (m)} & Baseline & 2218 & 2259 & 300\\
& ReWAG & 26 & 16 & 17\\\hline
\multirow{2}{*}{Final Standard Deviation (m)} & Baseline & 500 & 1321 & 169\\
& ReWAG & 18 & 10 & 10\\\hline
\multirow{2}{*}{Convergence Time (time steps)} & Baseline & - & - & -\\
& ReWAG & 133 & 61 & 41\\\hline
\end{tabular}
\end{center} \vspace*{-0.1in}
\end{table} 
% \multirow{2}{*}{Average Error (m)} & Baseline & 2241 & 2224 & 998\\
% & ReWAG & 925 & 591 & 2684\\\hline

\subsection{Small-scale Simulation: KITTI}
We also demonstrate ReWAG's localization performance on a simulated KITTI test path in Fig.~\ref{fig:kitti_path}. 
This experiment demonstrates the ability of ReWAG to localize with more accurate initialization across a smaller search area. We sample images and pose data from the residential ``2011{\_}0{\_}30{\_}drive{\_}0028'' path and reduce the FOV to 90\textdegree{} by cropping the KITTI images. Satellite images of the search area are obtained from the Google Static API; the search area is divided into a grid of 32 $\times$ 32 satellite images at zoom level 20. We initialize the particle filter with a Gaussian distribution centered roughly 80 m away from the true location, and after 34 ground images, ReWAG's final estimation error is 12 m.

\subsection{Large-scale Hardware Experiment: Boston} 
\begin{figure}[t!]
  \centering
  \subfigure[Close-up view of Boston paths.\label{fig:boston_zoom}]{\includegraphics[width=\columnwidth]{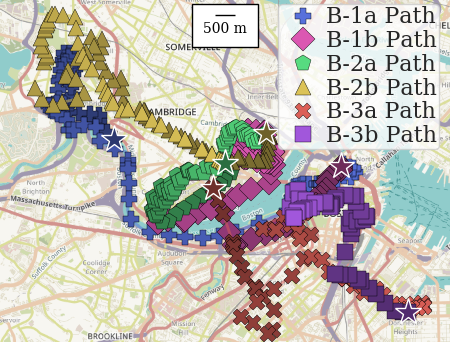}}
  \subfigure[Boston paths shown within full search area.\label{fig:boston}]{\includegraphics[width=\columnwidth]{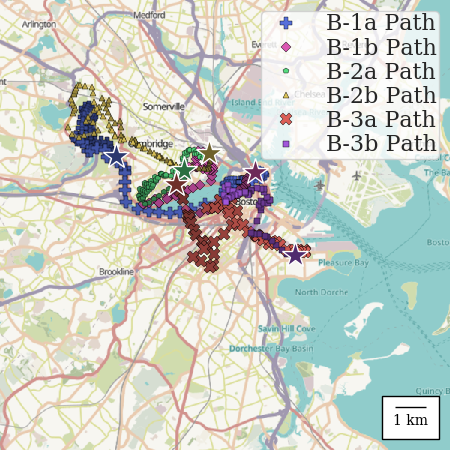}}
  \caption{Ground-truth paths in the Boston test area. Increasing brightness of path indicates passing of time. Start marked with stars. }
%   \vspace*{-0.2in}
  \label{fig:boston_paths}
\end{figure}

\begin{figure}[t!]
  \centering
  \subfigure[Photo of car with sensors.\label{fig:lincoln_photo}]{\includegraphics[width=.49\columnwidth]{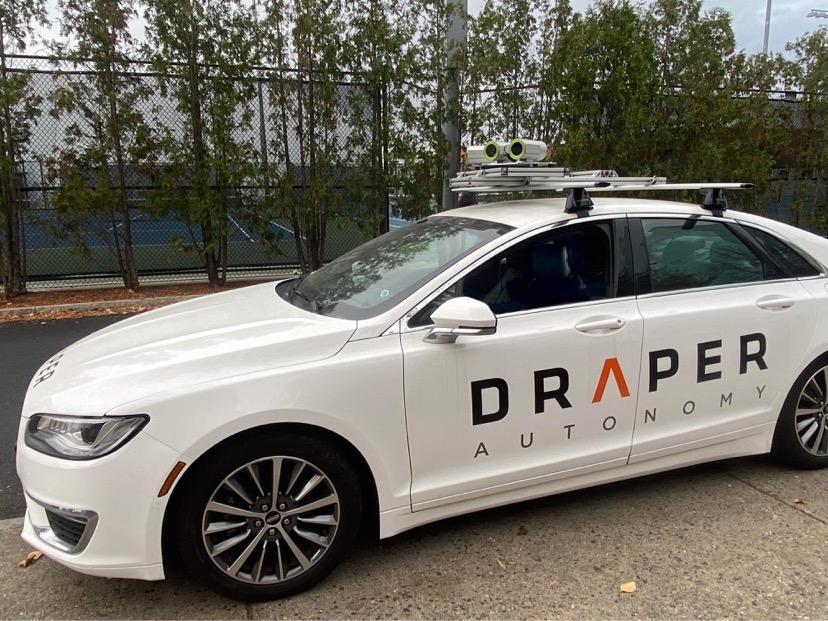}}
  \subfigure[System diagram. \label{fig:lincoln_sys}]{\includegraphics[width=.49\columnwidth]{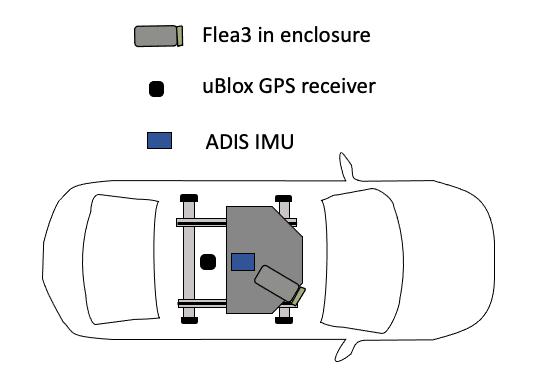}}
  \caption{Experimental setup of platform used for data collection.}
%   \vspace*{-0.2in}
  \label{fig:lincoln}
\end{figure}

We have collected approximately 6 hours of driving data in Boston, MA (and the adjacent city of Cambridge, MA) to further test ReWAG. This Boston data initiated changes which resulted in our evolved ReWAG, ReWAG*. The data was collected between November 11th, 2022 and November 21st, 2022 and is characterized in Table \ref{tab:boston_characteristics} with paths shown in Fig.~\ref{fig:boston_paths}. We drove a Lincoln MKZ (see Fig.~\ref{fig:lincoln}) equipped with a Point Grey Flea3 camera with Navitar 4.5 mm effective focal length lens, an ADIS 16448 IMU, and a uBlox GPS receiver. The camera was mounted within a weatherproof enclosure on a plate on top of the car, angled approximately 30$^\circ$ to the passenger side. The IMU was mounted on the underside of the plate within an EMI shielded enclosure, centered on the car. The GPS receiver was mounted on the roof of the car behind the camera plate. Our dataset consists of rectified RGB camera imagery taken approximately every 5 seconds along the driving paths and the corresponding GPS tags for those images. We obtained the GPS tags using Samwise \cite{samwise}, which fuses the GPS data from our uBlox receiver with visual-inertial odometry to obtain smoothed estimates of the precise location where each image was taken. We chose to angle our camera 30\textdegree{} to the right side of the vehicle because it shows more of the surrounding scenery instead of the road, which is less distinctive and may often be partially blocked with other cars. Satellite images were collected from Google Maps in the same manner as in the Chicago simulation. We initialized the particle filter with Gaussian distributions, centered 1.3 km from the true initial location (standard deviation of 500 m) for B-1b, B-2a, B-2b, B-3a and B-3b, and centered 800 m from the true initial location (standard deviation of 150 m) for B-1a.

\setlength\tabcolsep{1.5pt}
\begin{table}[t]
\vskip+0.1in
\caption{Test Path Characteristics}
\vspace*{-0.2in}
\label{tab:boston_characteristics}
% \centering
\begin{center}
\begin{tabular}{|l|c|c|c|c|c|}\hline
\bf Name & \bf Date &\bf Time of Day & \bf Weather & \bf Location& \bf Length\\\hline
B-1a & 11/11/22 & Afternoon  & Cloudy & Cambridge \& Boston & 15 km\\%\hline
B-1b & 11/11/22 & Afternoon  & Cloudy & Boston & 13 km\\%\hline
B-2a & 11/15/22 & Morning  & Sunny & Cambridge & 10 km\\%\hline
B-2b & 11/15/22 & Morning  & Partly Cloudy & Cambridge & 17 km \\%\hline
B-3a & 11/23/22 & Afternoon  & Sunny & Boston  & 14 km\\
B-3b & 11/23/22 & Afternoon  & Sunny & Boston  & 8 km\\\hline
\end {tabular}
% \centering
\end{center}
% \vspace*{-0.3in}
\end{table}

Localization on the images from this Boston dataset is a significant challenge for several reasons: ReWAG* was not trained on any images of Boston, the Boston images have a 72$^\circ$ FOV while ReWAG* was only trained on 90$^\circ$ FOV images, the lighting conditions vary significantly from training images, and the Boston images were taken at a different time of year than most training images (Fall vs.\ Summer). Despite these challenges, ReWAG* was able to localize to roughly the same level of accuracy in Boston as ReWAG did in the simulated Chicago data. 

\begin{figure}[t!]
  \centering
  \subfigure[Time t.\label{fig:charles_1}]{\includegraphics[width=.28\columnwidth]{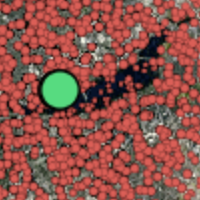}}
  \subfigure[Time t + 1.\label{fig:charles_2}]{\includegraphics[width=.28\columnwidth]{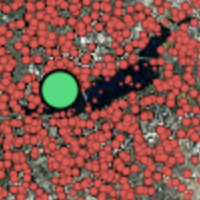}}
  \subfigure[Time t + 2.\label{fig:charles_3}]{\includegraphics[width=.28\columnwidth]{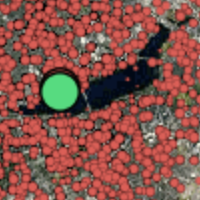}}
  \caption{Example of particle degeneration in Charles River. Red dots represent particles, green circle represents true location.}
%   \vspace*{-0.2in}
  \label{fig:charles}
\end{figure}

Our first tests on the Boston data revealed two key weaknesses of our initial version of ReWAG. First, ReWAG suffered from particle degeneracy that became particularly apparent in the B-1b and B3-b test paths, which spent significant time traveling on roads directly adjacent to the Charles River. Particles that were propagated into the river quickly degenerated and were never recovered, as illustrated in Fig.~\ref{fig:charles}. This made it unlikely for a path along the river to be accurately estimated, since noisy odometry makes it likely for particles on the roads adjacent to rivers to eventually be propagated into the river at some point in their trajectories, causing them to then degenerate. Our testing on these challenging paths prompted improvements in our particle filter resampling strategies for ReWAG*. 

\begin{figure}[t!]
  \centering
  \subfigure[Google StreetView image of Amherst Alley, Cambridge, MA. \label{fig:streetview_mit}]{\includegraphics[width=.48\columnwidth] {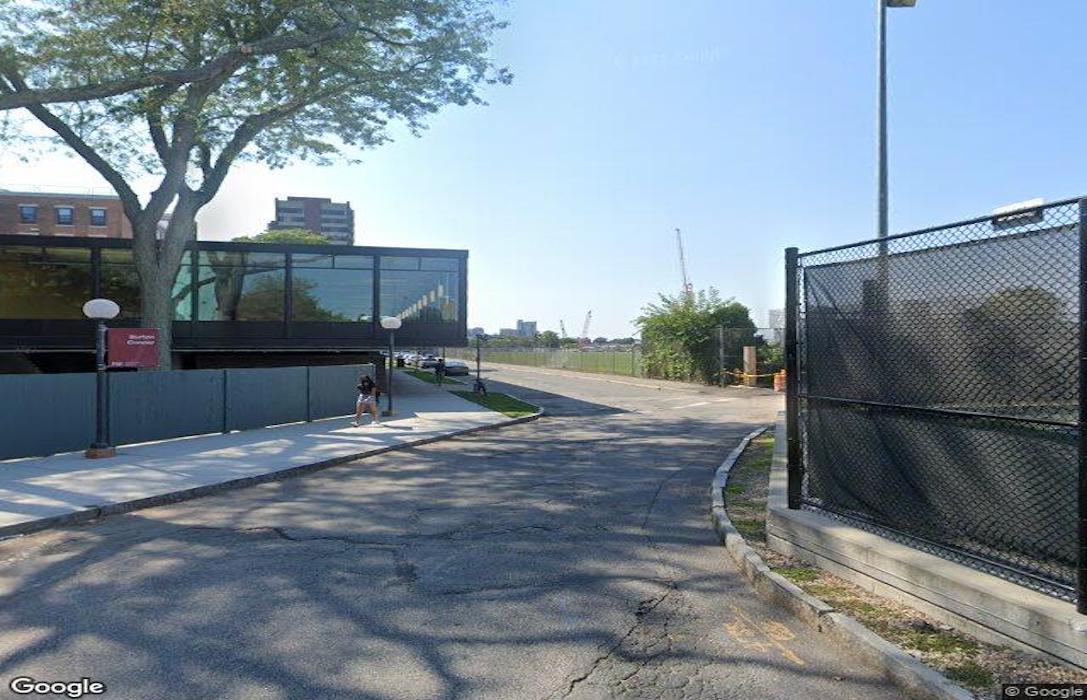}}
  \subfigure[Our collected image of Amherst Alley, Cambridge, MA on a cloudy day. \label{fig:draper_mit}]{\includegraphics[width=.48\columnwidth]  {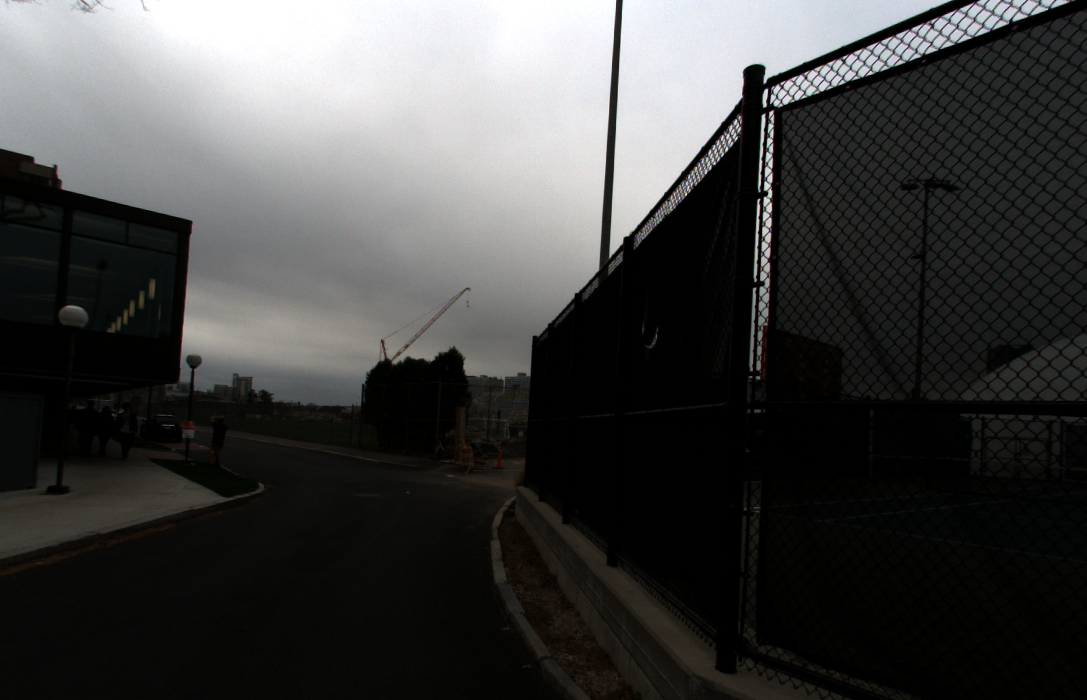}}
  \caption{Example of difference in appearance between Google StreetView image and self-collected image. Google StreetView images are carefully selected and processed, while in practice weather conditions may vary, and camera exposure and aperture may not be perfectly adjusted.}
%   \vspace*{-0.2in}
  \label{fig:streetview}
\end{figure}

Secondly, ReWAG was sensitive to the brightness of the ground images it was tested on. The B-1a and B1-b test paths, which were collected on an overcast, cloudy day, also highlighted this sensitivity. Originally, the Siamese network was trained on satellite images from Google Maps and ground images from Google StreetView without any data augmentation. However, Google StreetView images are not randomly dispersed in different weather conditions, they are overwhelmingly taken on bright and sunny, or partially sunny days, (see Fig.~\ref{fig:streetview}). 
Previously published Siamese networks for cross-view geolocalization had  not done data augmentation during training, so ReWAG  \cite{downes_icra_23} did not either. This lack of training on darker or lower exposure images made cloudy ground images out of the training distribution for ReWAG's Siamese network, and because ReWAG was not familiar with these kinds of images, it could not discriminate between matching ground-satellite pairs and non-matching pairs. This prompted the incorporation of training data augmentation into ReWAG*'s training.

ReWAG*'s results with both of these improvements can be seen in Table \ref{tab:summary_boston}, where ReWAG* outperforms the vision transformer baseline \cite{zhu2022transgeo} on every path.
\setlength\tabcolsep{2.5pt}
\begin{table}[t]
\caption{Boston Experiment Results}
\vspace*{-.2in}
\label{tab:summary_boston}
\begin{center}
\begin{tabular}{|cl|c|c|c|c|c|c|}
\hline
\multicolumn{2}{|c|}{\bf Metric and System Type}  & \bf B-1a  & \bf B-1b & \bf B-2a & \bf B-2b & \bf B-3a & \bf B-3b\\\hline
\multirow{2}{*}{Final Error (m)} & Baseline & 864 & 1329 & 1891 & 1884 & 1447 & 1690\\
& ReWAG* & 23 & 75 & 99 & 105 & 43 & 62\\\hline
\multirow{2}{*}{Conv. Time (t-steps)} & Baseline & - & - & - & - & - & -\\
& ReWAG* & 329 & 220 & 347 & 247 & 166 & 128\\\hline
\end{tabular}
\end{center} \vspace*{-0.1in}
\end{table}

\begin{figure}[t]
\centering
\includegraphics[width=\linewidth]{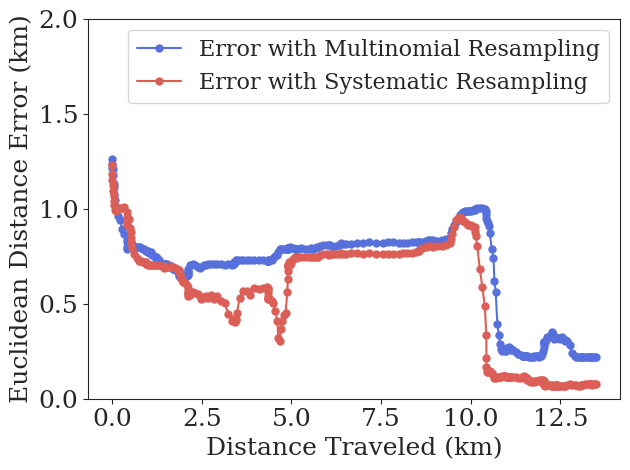}
\vspace*{-0.3in}
\caption{Comparison of ReWAG* with multinomial resampling and with systematic resampling on simulation of B1-b test path. Multinomial resampling does not converge, while systematic resampling converges after 10 km.}
  \label{fig:resample_abl} 
%\end{figure}
%
%\begin{figure}[t]
\centering
\includegraphics[width=\linewidth]{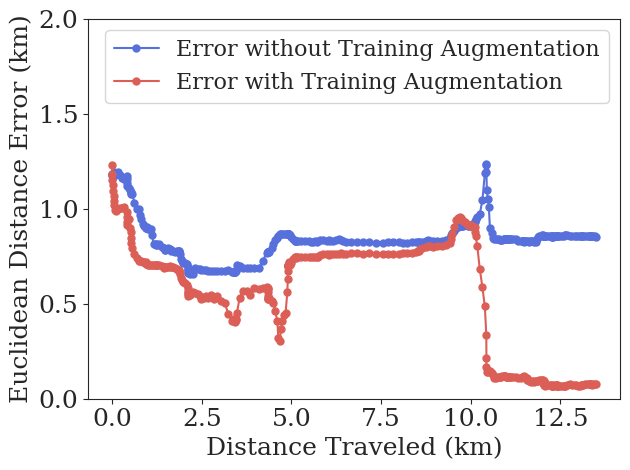}
\vspace*{-0.3in}
\caption{Comparison of ReWAG* performance on B-1b cloudy day with Fancy PCA training augmentation and without training augmentation.}
  \label{fig:augment_abl} 
\end{figure}

\textbf{Resampling ablation.} Fig.~\ref{fig:resample_abl} shows how, with the original multinomial resampling strategy from \cite{downes_icra_23}, ReWAG* fails to accurately converge on the challenging path along the Charles River. Multinomial resampling at every timestep (regardless of ESS) yields a final error of over 218 m. Systematic resampling with an 0.98 ESS was able to achieve a final error of 75 m and convergence after 220 timesteps. Table \ref{tab:summary_ablation_tro} summarizes the performance with multinomial resampling at every timestep compared to ReWAG* with systematic resampling only when ESS drops below 0.98. While multinomial resampling decreases final estimation error and time to convergence in some paths, in others it fails to converge at all. ReWAG* with systematic sampling converges on every test path and hence yields more reliable performance.

% \setlength\tabcolsep{2.5pt}
% \begin{table}[t]
% \caption{Resampling Ablation}
% \vspace*{-.2in}
% \label{tab:summary_resample_ablation}
% \begin{center}
% \begin{tabular}{|cl|c|c|c|c|c|c|}
% \hline
% \multicolumn{2}{|c|}{\bf Metric and System Type}  & \bf B-1a  & \bf B-1b & \bf B-2a & \bf B-2b & \bf B-3a & \bf B-3b\\\hline
% \multirow{2}{*}{Final Error (m)} & Multinomial & 23 & 218 & 75 & 167 & 60 & 223\\
% & Systematic & 23 & 75 & 99 & 105 & 43 & 62\\\hline
% \multirow{2}{*}{Conv. Time (t-steps)} & Multinomial & 254 & - & 347 & - & 136 & -\\
% & Systematic & 329 & 220 & 347 & 247 & 166 & 128\\\hline
% \end{tabular}
% \end{center} \vspace*{-0.1in}
% \end{table} 

\textbf{Data augmentation ablation.}
We encountered variable lighting conditions while collecting our data in Boston, including a dark, overcast day with diffuse, low lighting (in paths B-1a and B-1b) and a partly cloudy period with lower lighting (B-2b). These conditions were not seen frequently in the training dataset. To remedy this, we incorporated Fancy PCA \cite{krizhevsky2017imagenet} data augmentation into ReWAG*'s training to improve our Siamese network's robustness to brightness changes. We used $\alpha=1000$ in training to produce more drastic lighting differences. The results of training with Fancy PCA can be seen in Fig.~\ref{fig:augment_abl}, where ReWAG without Fancy PCA does not converge on the B-1b cloudy day and ReWAG with Fancy PCA does converge to 75 m of error. Table \ref{tab:summary_ablation_tro} summarizes the performance with, and without, Fancy PCA training data augmentation; showing that the data augmentation reduces final estimation error in all but the B-3b path, and especially reduces estimation error on low light paths.

% \setlength\tabcolsep{3pt}
% \begin{table}[t]
% \caption{Training Augmentation Ablation}
% \vspace*{-.2in}
% \label{tab:summary_augment_ablation}
% \begin{center}
% \begin{tabular}{|cl|c|c|c|c|c|c|}
% \hline
% \multicolumn{2}{|c|}{\bf Metric and System Type}  & \bf B-1a  & \bf B-1b & \bf B-2a & \bf B-2b & \bf B-3a & \bf B-3b\\\hline
% \multirow{2}{*}{Final Error (m)} & No Aug & 79 & 853 & 131 & 2206 & 125 & 61\\
% & Aug & 23 & 75 & 99 & 105 & 43 & 62\\\hline
% \multirow{2}{*}{Conv. Time (t-steps)} & No Aug & 302 & - & 382 & - & 168 & 87\\
% & Aug & 329 & 220 & 347 & 247 & 166 & 128\\\hline
% \end{tabular}
% \end{center} \vspace*{-0.1in}
% \end{table} 

\setlength\tabcolsep{1.9pt}
\begin{table}[t]
\caption{ReWAG* Ablation}
\vspace*{-.2in}
\label{tab:summary_ablation_tro}
\begin{center}
\begin{tabular}{|cl|c|c|c|c|c|c|}
\hline
\multicolumn{2}{|c|}{\bf Metric and System Type}  & \bf B-1a  & \bf B-1b & \bf B-2a & \bf B-2b & \bf B-3a & \bf B-3b\\\hline
\multirow{2}{*} & No Augmentation & 79 & 853 & 131 & 2206 & 125 & 61\\
{Final Error (m)} & Multinomial & 23 & 218 & 75 & 167 & 60 & 223\\
& ReWAG* & 23 & 75 & 99 & 105 & 43 & 62\\\hline
\multirow{2}{*} & No Augmentation & 302 & - & 382 & - & 168 & 87\\
{Conv. Time (t-steps)} & Multinomial & 254 & - & 347 & - & 136 & -\\
& ReWAG* & 329 & 220 & 347 & 247 & 166 & 128\\\hline
\end{tabular}
\end{center} \vspace*{-0.1in}
\end{table} 
\section{Conclusion}
ReWAG redesigns the Siamese network-particle filter architecture for increased hardware flexibility, a key component of applying cross-view geolocalization to mobile robotics. Previous works have largely focused on geolocalization with ground cameras that have much wider FOVs than cameras that would be readily available on existing platforms. This work uses all available information to generate Siamese network embeddings and accurately reweight particles, hence enabling faster and more accurate particle filter convergence with limited FOV cameras. 

ReWAG* generalizes from training on bright, 90\textdegree{} FOV images of Chicago to accurately localize across several hundred square kilometers of Boston with 72\textdegree{} FOV images with variable brightness. ReWAG also accurately localizes across several hundred square kilometers of Chicago in simulation, maintaining the same level of accuracy that \cite{downes_iros_22} previously demonstrated with panoramic ground cameras. ReWAG maintains the same benefits of WAG in terms of reducing the size of the satellite image database required. Additionally, it has lower final error and faster convergence than the TransGeo baseline in the Chicago test area, which we attribute to its training with trinomial loss. It also successfully converges to a location estimate that is 12 m from the true location in the KITTI test path, which further demonstrates its ability to generalize to a city it was not trained on.

In the short term, future work on this topic includes improving computational speed, testing on real-time physical platforms, and further restriction of the FOV. Domain shift remains an open challenge in the field---localizing in areas with different semantic appearances than training image pairs, on satellite images that were taken in different seasons than the ground images, and in rural areas without as many identifiable landmarks as residential or urban areas.

In summary, ReWAG lifts the heavy hardware requirements that were inherent in other cross-view geolocalization systems and enables accurate localization in previously unseen environments using a camera with as narrow a FOV as 72\textdegree{}. It represents a step forward in making cross-view geolocalization a widespread tool that can be used on many platforms with different sets of hardware. 

\balance
\bibliographystyle{IEEEtran}
\bibliography{root}

\newcommand{\noop}[1]{}
\begin{thebibliography}{10}
\providecommand{\url}[1]{#1}
\csname url@rmstyle\endcsname
\providecommand{\newblock}{\relax}
\providecommand{\bibinfo}[2]{#2}
\providecommand\BIBentrySTDinterwordspacing{\spaceskip=0pt\relax}
\providecommand\BIBentryALTinterwordstretchfactor{4}
\providecommand\BIBentryALTinterwordspacing{\spaceskip=\fontdimen2\font plus
\BIBentryALTinterwordstretchfactor\fontdimen3\font minus
  \fontdimen4\font\relax}
\providecommand\BIBforeignlanguage[2]{{%
\expandafter\ifx\csname l@#1\endcsname\relax
\typeout{** WARNING: IEEEtran.bst: No hyphenation pattern has been}%
\typeout{** loaded for the language `#1'. Using the pattern for}%
\typeout{** the default language instead.}%
\else
\language=\csname l@#1\endcsname
\fi
#2}}

\bibitem{Tian}
Y.~Tian, C.~Chen, and M.~Shah, ``Cross-view image matching for geo-localization
  in urban environments,'' in \emph{IEEE Conference on Computer Vision and
  Pattern Recognition (CVPR)}, 2017, pp. 1998--2006.

\bibitem{Shi}
Y.~Shi, L.~Liu, X.~Yu, and H.~Li, ``Spatial-aware feature aggregation for image
  based cross-view geo-localization,'' in \emph{NeurIPS}, 2019.

\bibitem{Cai}
S.~Cai, Y.~Guo, S.~Khan, J.~Hu, and G.~Wen, ``Ground-to-aerial image
  geo-localization with a hard exemplar reweighting triplet loss,'' in
  \emph{2019 IEEE/CVF International Conference on Computer Vision (ICCV)},
  2019, pp. 8390--8399.

\bibitem{Kim}
D.-K. Kim and M.~R. Walter, ``Satellite image-based localization via learned
  embeddings,'' \emph{2017 IEEE International Conference on Robotics and
  Automation (ICRA)}, pp. 2073--2080, 2017.

\bibitem{Hu}
\BIBentryALTinterwordspacing
S.~Hu and G.~H. Lee, ``Image-based geo-localization using satellite imagery,''
  \emph{International J.\ of Computer Vision}, pp. 1205--1219, 2020. [Online].
  Available: \url{https://doi.org/10.1007/s11263-019-01186-0}
\BIBentrySTDinterwordspacing

\bibitem{ZhuVIGOR}
S.~Zhu, T.~Yang, and C.~Chen, ``Vigor: Cross-view image geo-localization beyond
  one-to-one retrieval,'' \emph{2021 IEEE/CVF Conference on Computer Vision and
  Pattern Recognition (CVPR)}, pp. 5316--5325, 2021.

\bibitem{Liu2019}
L.~Liu and H.~Li, ``Lending orientation to neural networks for cross-view
  geo-localization,'' \emph{2019 IEEE/CVF Conference on Computer Vision and
  Pattern Recognition (CVPR)}, pp. 5617--5626, 2019.

\bibitem{Viswanathan}
A.~Viswanathan, B.~R. Pires, and D.~F. Huber, ``Vision based robot localization
  by ground to satellite matching in gps-denied situations,'' \emph{2014
  IEEE/RSJ International Conference on Intelligent Robots and Systems}, pp.
  192--198, 2014.

\bibitem{Zhu2021}
S.~Zhu, T.~Yang, and C.~Chen, ``Revisiting street-to-aerial view image
  geo-localization and orientation estimation,'' in \emph{2021 Winter Conf.\ on
  Applications of Computer Vision}, 01 2021, pp. 756--765.

\bibitem{Xia}
Z.~Xia, O.~Booij, M.~Manfredi, and J.~F.~P. Kooij, ``Cross-view matching for
  vehicle localization by learning geographically local representations,''
  \emph{IEEE Robotics and Automation Letters}, vol.~6, pp. 5921--5928, 2021.

\bibitem{downes_iros_22}
L.~M. Downes, D.-K. Kim, T.~J. Steiner, and J.~P. How, ``City-wide
  street-to-satellite image geolocalization of a mobile ground agent,'' in
  \emph{2022 IEEE/RSJ International Conference on Intelligent Robots and
  Systems (IROS)}, 2022, pp. 11\,102--11\,108.

\bibitem{zhu2022transgeo}
S.~Zhu, M.~Shah, and C.~Chen, ``Transgeo: Transformer is all you need for
  cross-view image geo-localization,'' in \emph{Proceedings of the IEEE/CVF
  Conference on Computer Vision and Pattern Recognition}, 2022, pp. 1162--1171.

\bibitem{downes_icra_23}
L.~M. Downes, T.~J. Steiner, R.~L. Russell, and J.~P. How, ``Wide-area
  geolocalization with a limited field of view camera,'' in \emph{2023 IEEE
  International Conference on Robotics and Automation (ICRA)}, 2023.

\bibitem{tibshirani1993introduction}
R.~J. Tibshirani and B.~Efron, ``An introduction to the bootstrap,''
  \emph{Monographs on statistics and applied probability}, vol.~57, no.~1,
  1993.

\bibitem{doucet2001sequential}
A.~Doucet, N.~De~Freitas, N.~J. Gordon, \emph{et~al.}, \emph{Sequential Monte
  Carlo methods in practice}.\hskip 1em plus 0.5em minus 0.4em\relax Springer,
  2001, vol.~1, no.~2.

\bibitem{liu1996metropolized}
J.~S. Liu, ``Metropolized independent sampling with comparisons to rejection
  sampling and importance sampling,'' \emph{Statistics and computing}, vol.~6,
  pp. 113--119, 1996.

\bibitem{krizhevsky2017imagenet}
A.~Krizhevsky, I.~Sutskever, and G.~E. Hinton, ``Imagenet classification with
  deep convolutional neural networks,'' \emph{Communications of the ACM},
  vol.~60, no.~6, pp. 84--90, 2017.

\bibitem{viswanathan2016}
A.~Viswanathan, B.~R. Pires, and D.~Huber, ``Vision-based robot localization
  across seasons and in remote locations,'' in \emph{2016 IEEE International
  Conference on Robotics and Automation (ICRA)}, 2016, pp. 4815--4821.

\bibitem{jacobs}
N.~Jacobs, S.~Satkin, N.~Roman, R.~Speyer, and R.~Pless, ``Geolocating static
  cameras,'' in \emph{2007 IEEE 11th International Conference on Computer
  Vision}, 2007, pp. 1--6.

\bibitem{bansal}
\BIBentryALTinterwordspacing
M.~Bansal, H.~S. Sawhney, H.~Cheng, and K.~Daniilidis, ``Geo-localization of
  street views with aerial image databases,'' in \emph{Proceedings of the 19th
  ACM International Conference on Multimedia}, ser. MM '11.\hskip 1em plus
  0.5em minus 0.4em\relax New York, NY, USA: Association for Computing
  Machinery, 2011, p. 1125–1128. [Online]. Available:
  \url{https://doi-org.libproxy.mit.edu/10.1145/2072298.2071954}
\BIBentrySTDinterwordspacing

\bibitem{Workman}
S.~Workman, R.~Souvenir, and N.~Jacobs, ``Wide-area image geolocalization with
  aerial reference imagery,'' \emph{2015 IEEE International Conference on
  Computer Vision (ICCV)}, pp. 3961--3969, 2015.

\bibitem{Vo}
N.~N. Vo and J.~Hays, ``Localizing and orienting street views using overhead
  imagery,'' in \emph{ECCV}, 2016.

\bibitem{Rodrigues}
\BIBentryALTinterwordspacing
R.~Rodrigues and M.~Tani, ``Are these from the same place? seeing the unseen in
  cross-view image geo-localization,'' in \emph{2021 IEEE Winter Conference on
  Applications of Computer Vision (WACV)}.\hskip 1em plus 0.5em minus
  0.4em\relax Los Alamitos, CA, USA: IEEE Computer Society, jan 2021, pp.
  3752--3760. [Online]. Available:
  \url{https://doi.ieeecomputersociety.org/10.1109/WACV48630.2021.00380}
\BIBentrySTDinterwordspacing

\bibitem{Cao}
R.~Cao, J.~Zhu, Q.~Li, Q.~Zhang, Q.~Li, B.~Liu, and G.~Qiu, ``Learning
  spatial-aware cross-view embeddings for ground-to-aerial geolocalization,''
  in \emph{ICIG}, 2019.

\bibitem{lin}
T.-Y. Lin, S.~Belongie, and J.~Hays, ``Cross-view image geolocalization,'' in
  \emph{2013 IEEE Conference on Computer Vision and Pattern Recognition}, 2013,
  pp. 891--898.

\bibitem{lin2015}
T.-Y. Lin, Y.~Cui, S.~Belongie, and J.~Hays, ``Learning deep representations
  for ground-to-aerial geolocalization,'' in \emph{2015 IEEE Conference on
  Computer Vision and Pattern Recognition (CVPR)}, 2015, pp. 5007--5015.

\bibitem{bromley}
J.~Bromley, J.~Bentz, L.~Bottou, I.~Guyon, Y.~Lecun, C.~Moore, E.~Sackinger,
  and R.~Shah, ``Signature verification using a "siamese" time delay neural
  network,'' \emph{International Journal of Pattern Recognition and Artificial
  Intelligence}, vol.~7, p.~25, 08 1993.

\bibitem{shi2020looking}
Y.~Shi, X.~Yu, D.~Campbell, and H.~Li, ``Where am i looking at? joint location
  and orientation estimation by cross-view matching,'' in \emph{Proceedings of
  the IEEE/CVF Conference on Computer Vision and Pattern Recognition}, 2020,
  pp. 4064--4072.

\bibitem{rodrigues2022global}
R.~Rodrigues and M.~Tani, ``Global assists local: Effective aerial
  representations for field of view constrained image geo-localization,'' in
  \emph{Proceedings of the IEEE/CVF Winter Conference on Applications of
  Computer Vision}, 2022, pp. 3871--3879.

\bibitem{shi2022beyond}
Y.~Shi and H.~Li, ``Beyond cross-view image retrieval: Highly accurate vehicle
  localization using satellite image,'' in \emph{Proceedings of the IEEE/CVF
  Conference on Computer Vision and Pattern Recognition}, 2022, pp.
  17\,010--17\,020.

\bibitem{vgg}
K.~Simonyan and A.~Zisserman, ``Very deep convolutional networks for
  large-scale image recognition,'' in \emph{International Conference on
  Learning Representations}, 2015.

\bibitem{dash_review_2022}
\BIBentryALTinterwordspacing
T.~Dash, S.~Chitlangia, A.~Ahuja, and A.~Srinivasan, ``A review of some
  techniques for inclusion of domain-knowledge into deep neural networks,''
  \emph{Scientific Reports}, vol.~12, no.~1, p. 1040, Jan. 2022. [Online].
  Available: \url{https://doi.org/10.1038/s41598-021-04590-0}
\BIBentrySTDinterwordspacing

\bibitem{hol}
J.~D. Hol, T.~B. Schon, and F.~Gustafsson, ``On resampling algorithms for
  particle filters,'' in \emph{2006 IEEE Nonlinear Statistical Signal
  Processing Workshop}, 2006, pp. 79--82.

\bibitem{Ristic2004BeyondTK}
B.~Ristic, S.~Arulampalam, and N.~J. Gordon, \emph{Beyond the Kalman Filter:
  Particle Filters for Tracking Applications}.\hskip 1em plus 0.5em minus
  0.4em\relax Boston, MA: Artech House, 2004.

\bibitem{KUPTAMETEE2022110836}
\BIBentryALTinterwordspacing
C.~Kuptametee and N.~Aunsri, ``A review of resampling techniques in particle
  filtering framework,'' \emph{Measurement}, vol. 193, p. 110836, 2022.
  [Online]. Available:
  \url{https://www.sciencedirect.com/science/article/pii/S0263224122001312}
\BIBentrySTDinterwordspacing

\bibitem{samwise}
T.~J. Steiner, R.~D. Truax, and K.~Frey, ``A vision-aided inertial navigation
  system for agile high-speed flight in unmapped environments: Distribution
  statement a: Approved for public release, distribution unlimited,'' in
  \emph{2017 IEEE Aerospace Conference}, 2017, pp. 1--10.

\end{thebibliography}

\end{document}